\documentclass[letterpaper, 10 pt, conference]{ieeeconf}  

\IEEEoverridecommandlockouts                              

\usepackage[
    colorlinks,
    linkcolor=red,
    anchorcolor=blue,
    citecolor=green
]{hyperref}

\usepackage{epsfig} 
\usepackage{mathptmx} 
\usepackage{times} 
\usepackage{graphicx}
\usepackage{amsmath} 
\usepackage{amssymb}  
\usepackage{subcaption}
\usepackage{booktabs}
\usepackage{cleveref}
\usepackage{array}
\usepackage{multirow}
\usepackage{cite}

\newcommand{\fireEmoji}{\includegraphics[height=0.7em,trim=0 .4em 0 0]{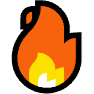}\hspace{.3em}}
\newcommand{\iceEmoji}{\includegraphics[height=0.7em,trim=0 .4em 0 0]{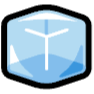}\hspace{.3em}}

\title{\LARGE \bf
Depth Helps: Improving Pre-trained RGB-based Policy with 

Depth Information Injection
}

\author{Xincheng Pang$^{1,2,\ast}$,  Wenke Xia$^{1,2,\ast}$, Zhigang Wang$^{2}$, Bin Zhao$^{2,3}$, Di Hu$^{1,\dagger}$, Dong Wang$^{2,\dagger}$, Xuelong Li$^{2,4}$
\thanks{$^{1}$Gaoling School of Artificial Intelligence, Renmin University of China}%
\thanks{$^{2}$Shanghai Artificial Intelligence Laboratory}%
\thanks{$^{3}$Northwestern Polytechnical University}%
\thanks{$^{4}$Institute of Artificial Intelligence, China Telecom Corp Ltd}%
\thanks{$^{\ast}$Equal contribution. Work is done during internship at Shanghai Artificial Intelligence Laboratory}%
\thanks{$^{\dagger}$Corresponding author}%
}

\begin{document}

\maketitle
\thispagestyle{empty}
\pagestyle{empty}

\begin{abstract}
3D perception ability is crucial for generalizable robotic manipulation. While recent foundation models have made significant strides in perception and decision-making with RGB-based input, their lack of 3D perception limits their effectiveness in fine-grained robotic manipulation tasks. 
To address these limitations, we propose a Depth Information Injection ($\bold{DI}^{\bold{2}}$) framework that leverages the RGB-Depth modality for policy fine-tuning, while relying solely on RGB images for robust and efficient deployment. Concretely, we introduce the Depth Completion Module (DCM) to extract the spatial prior knowledge related to depth information and generate virtual depth information from RGB inputs to aid policy deployment. Further, we propose the Depth-Aware Codebook (DAC) to eliminate noise and reduce the cumulative error from the depth prediction. In the inference phase, this framework employs RGB inputs and accurately predicted depth data to generate the manipulation action. We conduct experiments on simulated LIBERO environments and real-world scenarios, and the experiment results prove that our method could effectively enhance the pre-trained RGB-based policy with 3D perception ability for robotic manipulation. The website is released at \href{https://gewu-lab.github.io/DepthHelps-IROS2024}{https://gewu-lab.github.io/DepthHelps-IROS2024}.

\end{abstract}

\section{INTRODUCTION}

Building a generalizable manipulation policy is essential for the development of intelligent robotics.  
Foundation models have achieved remarkable success in various decision-making problems~\cite{wei2022chain}. Leveraging these models offers a promising approach to developing generalizable manipulation capabilities in robotics.
To deploy foundation models for generalizable robotic manipulation, recent research has taken different approaches. Some works use the world knowledge embedded in foundation models for instruction decomposition~\cite{lin2023text2motion}. Others focus on generating generalizable manipulation policies by leveraging extensive robotic datasets~\cite{brohan2023rt,octo_2023}.
Despite impressive achievements in manipulation tasks, relying solely on RGB perception limits the understanding of 3D environments. This constraint hinders robotic performance in fine-grained manipulation tasks.
In response, this work targets to utilize minimal aligned RGB-D data to enhance the 3D scene perception for robotic manipulation in wide-spread RGB-only scenarios.

To enhance unimodal perception through multimodal integration, two prominent strategies have emerged. These are cross-modal knowledge distillation~\cite{tian2019contrastive,yuhong-CMKD-ECCV2022} and missing modality learning~\cite{parthasarathy2020training,lee2023multimodal}.
The former approach uses a multi-modal teacher to distill cross-modal knowledge into an unimodal model. The latter method employs data augmentation across different modalities to encourage robust model learning.
Despite the success of these methods in various perception tasks~\cite{chen2021distilling, Guo_2021_ICCV}, their application to fine-grained robotic manipulation policy learning is challenging. This difficulty arises from sequential accumulative errors in multimodal perception prediction.

\begin{figure}[t]
      \centering
      \resizebox{0.8\linewidth}{!}{\includegraphics{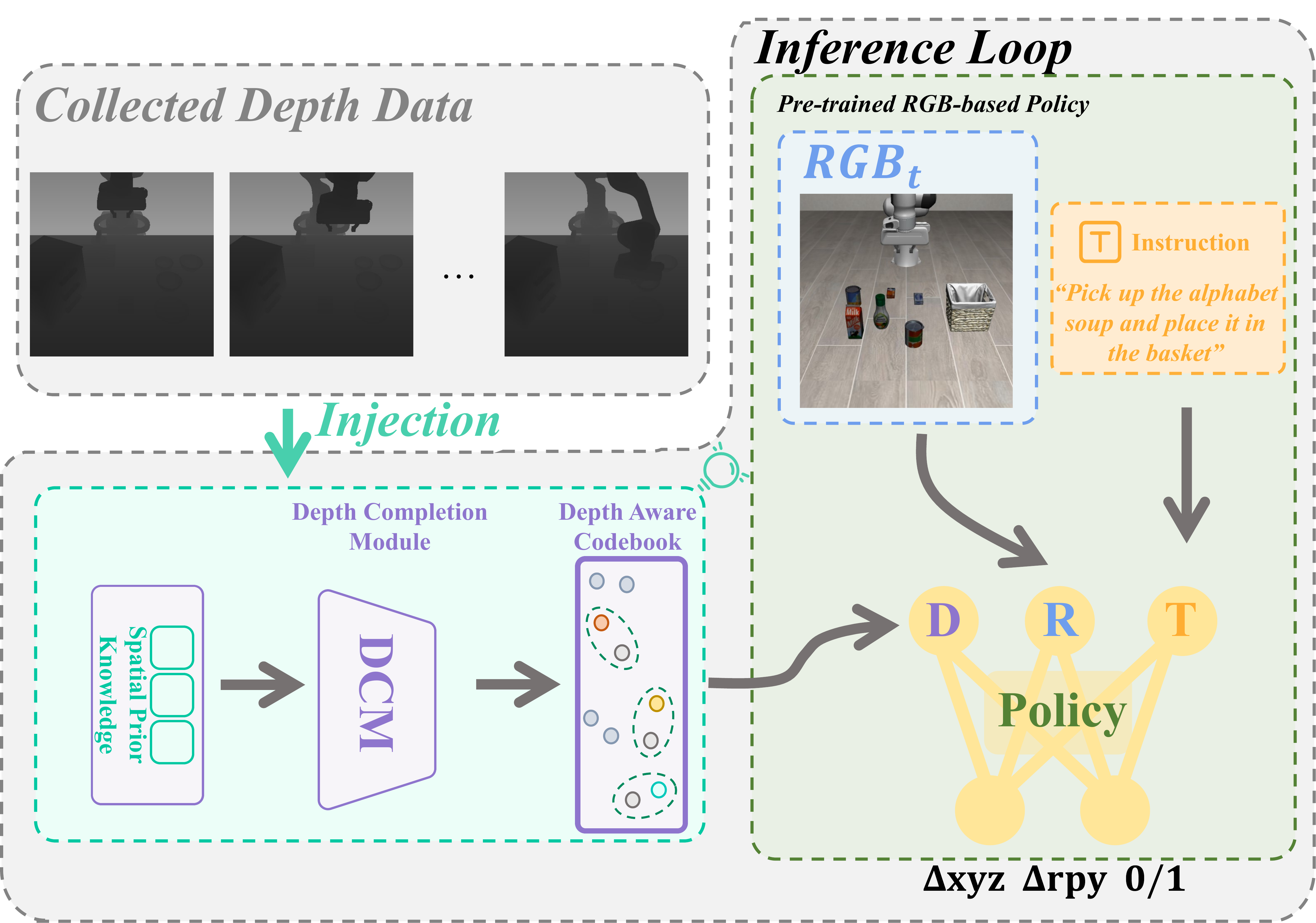}}
      \caption{
      We propose the Depth Information Injection framework to inject the spatial prior knowledge from the depth information into the RGB-based policy.
      }
      \label{figurelabel}
      \vspace{-0.7cm}
\end{figure}

To address this challenge and improve the 3D perception capabilities of pre-trained models for fine-grained manipulation tasks, we propose the Depth Information Injection (${\rm DI}^2$) framework. This framework includes the Depth Completion Module (DCM) for integrating spatial prior knowledge with RGB images, and the Depth-Aware Codebook (DAC) to reduce cumulative errors.
Specifically, we introduce the DCM to predict depth information from RGB inputs for use in complex environments. This module utilizes the Perceiver Resampler~\cite{alayrac2022flamingo} to capture modality-specific biases related to depth, ensuring accurate predictions. 
We further propose the DAC to refine depth prediction. The DAC transforms input features into discrete quantized vectors, emphasizing crucial information and filtering out modality-specific noise.
By aligning quantized vectors from RGB-predicted depth features with those from auxiliary depth data, we accurately represent depth for policy deployment. 
During inference, the fine-tuned policy uses RGB inputs along with quantized depth features to generate actions. This approach enhances execution efficacy and reliability in complex scenarios.

To validate the effectiveness of our framework, we conducted experiments on the LIBERO~\cite{liu2024libero} benchmark. The experimental results demonstrate excellent performance across a wide range of tasks. Furthermore, we also deploy our method in real-world scenarios, which proves its effectiveness and reliability in practical applications.

Our main contributions can be summarized as follows:
\begin{itemize}
    \item We propose the ${\rm DI}^2$ framework, which enhances the 3D perception ability of pre-trained RGB-based policy.
    \item We design the Depth Completion Module and Depth-Aware Codebook to extract modality-specific knowledge for depth prediction on policy deployment.
    \item Experiments on the LIBERO benchmark and in real-world scenarios validate the effectiveness of our method. We also demonstrate its potential for application in fine-grained manipulation tasks.
\end{itemize}

\section{RELATED WORK}

\subsection{Robotic Foundation Models}

Foundation models have achieved significant success in perception and decision-making tasks. These include visual grounding~\cite{wu2023next}, visual question answering~\cite{alayrac2022flamingo,awadalla2023openflamingo}, and more~\cite{wei2022learning}. 
Recent works~\cite{xia2023kinematic,brohan2023rt,octo_2023} have been inspired by the rich world knowledge inherent in foundation models. They aim to equip these models with embodied agents for interactive environments.
Early works~\cite{wang2023voyager,lin2023text2motion} focused on task planning and decomposed complex instructions into sub-goals with a pre-defined skill library.
To leverage foundation models for robotic action control, recent work~\cite{octo_2023} designed an efficient language-conditioned manipulation policy using extensive robotic data. Other research~\cite{li2023vision} focused on integrating Visual-Language Models (VLM) to utilize their broad world knowledge for robotic manipulation tasks.
Despite notable successes in manipulation tasks, the exclusive reliance on RGB perception has limited the potential. 
In this work, we propose an effective method to enhance the 3D perception ability of the pre-trained manipulation models. 

\subsection{Cross-Modal Knowledge Distillation}

Cross-modal knowledge distillation is crucial for transferring knowledge between modalities, enhancing the representational capabilities of a target modality. 
Previous work~\cite{tian2019contrastive,yuhong-CMKD-ECCV2022} achieved feature-level knowledge transfer using various loss functions. Other studies~\cite{chong2022monodistill} developed feature imitation methods that enhance scene perception by integrating knowledge from monocular cameras and depth sensors.
Beyond these, some work~\cite{chen2021distilling,xia2023robust} proposed the adaptive transferring method for robust and effective cross-modal knowledge distillation. 
However, there is still little research in the field of robotics. 
Our method addresses the need for extensive robotic RGB-D data by injecting prior depth knowledge into a pre-trained policy. This approach enhances 3D perception for precise manipulation while using minimal data.

\subsection{Missing Modality Learning}

Missing modality learning enhances model performance in scenarios with incomplete inputs. Previous work~\cite{parthasarathy2020training,lee2023multimodal} used data augmentation to train models for robust performance in missing modality scenarios, while others~\cite{zhao2021missing,ma2021smil,cai2018deep,YANG2024110082} predicted missing modalities during inference using an auxiliary model.
In this work, we predict missing modalities with a depth-aware codebook to discretize the predicted features. This method enhances prediction robustness and ensures trajectory accuracy in decision-making.

\section{METHOD}\label{sec:method}

\begin{figure*}[htpb]
  \centering
  \begin{minipage}{\textwidth}
        \centering
        \resizebox{0.8\linewidth}{!}{\includegraphics{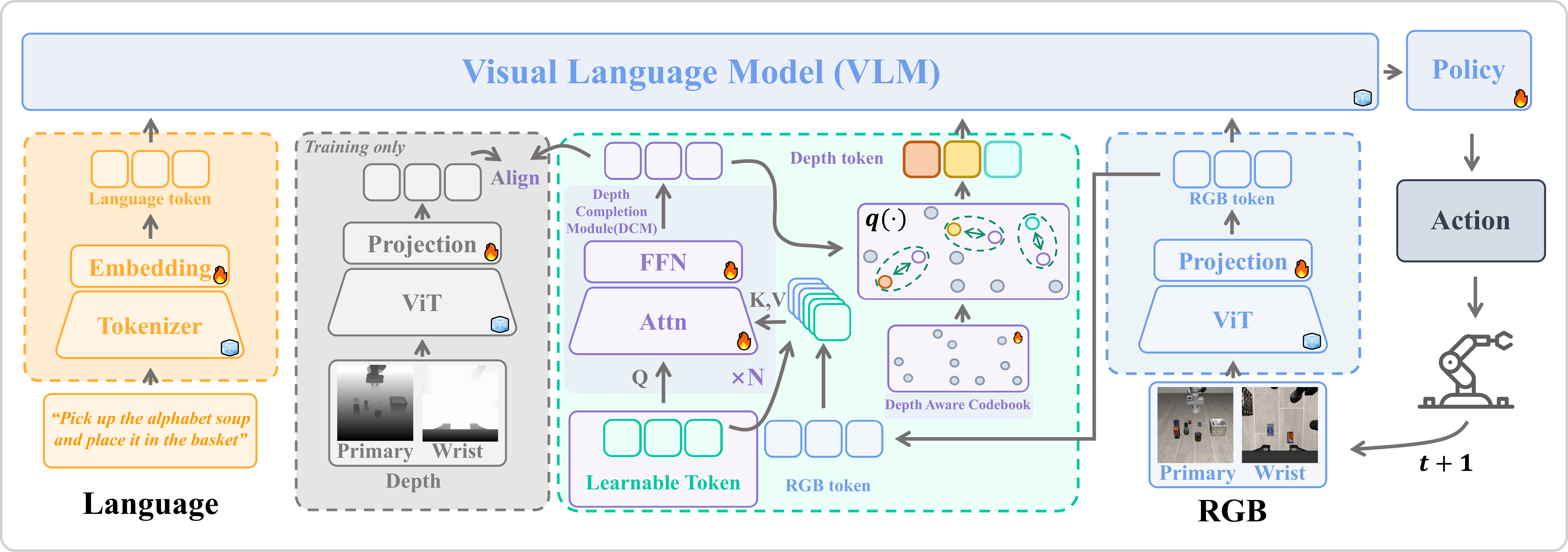}}

        \caption{Overview of our framework. \fireEmoji denotes the model parameters are updated, while \iceEmoji indicates the model parameters are frozen. During training, we use the collected depth image to train the Depth Completion Module and Depth-Aware Codebook. During inference, we use the Depth Completion Module together with the RGB token to predict the depth token.}
        \label{fig:overview_method}
    \end{minipage}
    \vspace{-0.7cm}
\end{figure*}

In this work, we introduce the ${\rm DI}^2$ framework. It uses RGB-Depth data for policy fine-tuning but relies solely on RGB images for robust and efficient deployment.
This framework comprises two core modules: the Depth Completion Module and the Depth-Aware Codebook.

\subsection{Overview}
The overall framework of the model is shown in \Cref{fig:overview_method}. When the model takes RGB-D as input during training, the entire model can be expressed as follows:
\begin{small} \begin{equation}
    \begin{aligned}
        a_t &= \pi \left({\rm VLM}\left(f^{text},f_t^{rgb},f_t^{depth}\right)\right), \\
    \end{aligned}
    \label{eq:rgbd_overview}
\end{equation} \end{small}
where $\pi$ is the policy model to get the final action. We utilize the VLM to extract features. The $f^{text}$, $f_t^{rgb}$ and $f_t^{depth}$ are extracted feature.

The RGB-D model, as detailed in \Cref{eq:rgbd_overview}, effectively utilizes depth images for 3D perception but is limited by its reliance on depth images. To overcome this limitation and enable manipulation in prevalent RGB-only scenarios, we introduce the Depth Completion Module (DCM). This module predicts the depth feature $\hat{f}_t^{depth}$ from the RGB image feature $f_t^{rgb}$, incorporating spatial prior knowledge $P$:
\begin{small} \begin{equation}
    \begin{aligned}
        \hat{f}_{t}^{depth} = {\rm DCM}\left(f_t^{rgb}, P\right).
    \end{aligned}
\end{equation} \end{small}
To reduce the cumulative errors, we further propose a Depth-Aware Codebook to discretize the depth features predicted by the DCM:
\begin{small} \begin{equation}
    \begin{aligned}
        \widetilde{f}_{t}^{depth} = {\rm Codebook}\left(\hat{f}_{t}^{depth}\right).
    \end{aligned}
\end{equation} \end{small}

When we only have RGB as the input during inference, we input the $f^{text}$, $f_t^{rgb}$, and $\widetilde{f}_{t}^{depth}$ together into the VLM to extract features. Ultimately, these features are fed into the policy model $\pi$ to get the final action.

\subsection{Depth Completion Module}
To enable the model to make accurate decisions with only RGB as input and leverage the spatial perception capabilities provided by the depth images during training, we introduce the Depth Completion Module (DCM). 
This module helps the model extract spatial information from RGB features without depth data and use it in decision-making.

To train the DCM, we utilize the collected trajectory data with RGB-D modalities. 
Specifically, for each training sample, we use a frozen visual encoder ViT and a specific projection layer ${\rm Proj}^{rgb}$ to extract the feature of RGB image $f^{rgb}_t = {\rm Proj}^{rgb}(ViT(o^{rgb}_t))$. 
As for the depth image, we make the shape of the depth image $o^{depth}_t$ the same as the RGB image $o^{rgb}_t$, and then put it into the same frozen visual encoder ViT to extract features. 
The difference lies in that we use a separate projection layer ${\rm Proj}^{depth}$ to compress and filter the depth image features $ViT(o^{depth}_t)$. 
Then we can obtain the filtered depth feature representation $f^{depth}_t = {\rm Proj}^{depth}(ViT(o^{depth}_t))$. 
Inspired by Perceiver Resampler~\cite{alayrac2022flamingo}, we integrate $k$ learnable tokens $P \in \mathbb{R}^{k \times d}$ into the DCM to explicitly capture the spatial prior knowledge. 
Based on the current timestep's RGB image feature $f^{rgb}_t$ and these learnable tokens $P$, the DCM estimates the depth feature $\hat{f}^{depth}_t$. 
In our approach, the parameter $P$ is designated as the query. We concatenate the RGB features $f^{rgb}_{t}$, with $P$ to construct both the key and the value components. These are then subjected to processing via cross-attention layers to enhance the feature integration.
In essence, $P$ stores spatial prior knowledge based on the similarity between training scenarios and testing scenarios. This enables us to leverage the statistical characteristics within $P$ to predict the depth features related to manipulation. Then the output will be passed into the FFN Layer to get the final feature. The above process can be described as follows:
\begin{small} \begin{equation}
    \begin{aligned}
        \tilde{Q}_{i+1} &= {\rm Attn}\left(Q_{i}, \left[Q_{i}, \left|\right.f^{rgb}_{t}\right], \left[Q_{i}, \left|\right.f^{rgb}_{t}\right]\right), \\ 
        Q_{i+1} &= {\rm FFN}\left(\tilde{Q}_{i+1}\right),
    \end{aligned}
\end{equation} \end{small}
where $i \in \left[0, N\right)$ means the i-th layer. $P$ is used as $Q_0$, $Q_N$ is seen as the estimate depth feature $\hat{f}^{depth}_{t}$.

\subsection{Depth-Aware Codebook}
The DCM can obtain an approximate representation of depth features, but there are still errors. 
Since manipulation involves temporal predictions, errors accumulate over time. This leads to a growing discrepancy between action sequences generated by the depth completion module and those using depth ground truth.
To address this issue, we introduce the Depth-Aware Codebook (DAC).

Mathematically, we define $Z \in \mathbb{R}^{N \times d}$ as the codebook, where $N$ represents the size of the codebook. Given a depth feature $f^{depth} \in \mathbb{R}^{n_{d} \times d}$, we utilize the quantization operation $\mathbf{q}\left(\cdot\left|Z\right.\right)$ to search for the $n_{d}$ closest vectors in $Z$ to achieve the discretization of $\hat{f}^{depth}$:
\begin{small} \begin{equation}
    \small
    \hat{f}^{depth}_{t} = \mathbf{q}\left(f^{depth}_{t}\left|Z\right.\right) = z_{k},\ \text{where}\ 
    k
    = \underset{z_k\in Z}{\operatorname{argmin}}\lVert f^{depth}_{t}-e_k\rVert.
\end{equation} \end{small}

Intuitively, the codebook functions as a set of clustering centroids for the depth features in the training dataset, offering higher robustness compared to the original depth features. This characteristic makes the model more reliable when dealing with noise and variations in practical applications.

\subsection{Training Details}

\textbf{Warm-up.}
We conduct a warm-up training process for manipulation with RGB and depth inputs in the collected trajectories. 
In this step, we directly train an imitation learning policy with the joint of depth model and frozen pre-trained RGB-based model as shown in \Cref{eq:warmup_LOSS}.
\begin{small} \begin{equation}
    \begin{aligned}
        L_{warmup} = \sum \limits_{t} \left|\left|\pi\left(\hat{a}_{t}|o^{rgb}_t, o^{depth}_t, L\right) - a_t\right|\right|^2.
    \end{aligned} 
    \label{eq:warmup_LOSS}
\end{equation} \end{small}

\textbf{Align.}
To remove the dependency of depth images, we further train the DCM using the perceptual encoder obtained from the warm-up phase. By utilizing paired RGB-D data in the collected trajectories, our objective is to inject 3D perception knowledge into the policy. We use \Cref{eq:dcm_loss} as the loss to train it.
\begin{small} \begin{equation}
    \begin{aligned}
        L_{dcm} &= \sum \limits_{t} \left|\left| DCM\left(P, {\rm sg}\left(f^{rgb}_{t}\right) \right)
- {\rm sg}\left(f^{depth}_{t}\right)\right|\right|^2, \\
    \end{aligned}
    \label{eq:dcm_loss}
\end{equation} \end{small}
where ${\rm sg}\left(\cdot\right)$ means the stop gradient operation, $f^{\cdot}_{t}$ is obtained by the encoder trained in the warm-up phase.

\textbf{Codebook.}
This phase is independent of the Align phase. We freeze the depth branch obtained in the warm-up phase and only train the Codebook with MSE loss. However, if we only utilize this loss, it could lead to the codebook collapse problem\cite{takida2022sq}. Inspired by CVQ-VAE\cite{Zheng_2023_CVQ}, we reinitialize the unoptimized points in each iteration. 

Specifically, in each iteration, we perform a running average operation for each codeword $c_k$ in the codebook:
\begin{small} \begin{equation}
    \begin{aligned}
        p_k &= p_k \lambda + \frac{n_k}{B}\left(1 - \lambda\right), \\
        \alpha_k &= \exp{\left(-p_k \frac{10N}{1 - \lambda}\right)}, \\
        c_k &= c_k \left(1 - \alpha_k\right) + \Bar{z}_k \alpha_k, \\
    \end{aligned}
\end{equation} \end{small}
where the $n_k$ represents the number of times $c_k$ is used in this iteration. $B$ means the batch size. $\lambda$ is a hyper-parameter. $\Bar{z}_k$ is an anchor vector that is chosen by probabilistic sampling. 
In our experiments, we set $N = 512, \lambda = 0.99$.

\section{SIMULATION EXPERIMENTS}\label{sec:experiments}
\subsection{Experiment Setup}
To validate the effectiveness of our proposed method, we conduct a series of experiments on the LIBERO benchmark~\cite{liu2024libero}. 
LIBERO is a large-scale benchmark with 130 robot manipulation tasks. It emphasizes scene diversity and task complexity, enabling fair comparisons and providing comprehensive evaluation results.
LIBERO is divided into four sub-suites: Object, Spatial, LongHorizon, and Goal. Each focuses on different manipulation skills, including object interaction, spatial perception, long-term decision-making, and goal-oriented tasks.
In terms of training data, we utilize the 50 high-quality human teleoperation demonstration trajectories provided by LIBERO for each task, totaling 6,500 trajectories. 
We train a multi-task model using these data and evaluate it on the four suites. Regarding the model architecture, we adopt RoboFlamingo~\cite{li2023vision} pre-trained in Calvin~\cite{mees2022calvin} as our model. 

\subsection{Preliminary Experiment}\label{sec:preliminary_experiment}
To get the upper bound of our method, we first conduct a preliminary experiment where the model could obtain both RGB images and depth images as input. We compare the following methods:
\begin{itemize}
    \item \textbf{RGB-RF}~\cite{li2023vision}: The standard RoboFlamingo architecture. 
    This baseline uses an LLM to extract text features from task instructions and a ViT to extract visual features from RGB images. These features are then fused using OpenFlamingo~\cite{awadalla2023openflamingo}.
    Finally, the fused features are input into the policy network to predict action.
    \item \textbf{RGB-D-RF}: Based on RGB-RF, we add an extra branch to extract features from depth images. 
    We concatenate RGB and depth image features along the channel dimension, then apply the same method as RGB-RF to map the fused features to actions.
    \item \textbf{Data Aug}~\cite{parthasarathy2020training}: This method augments the training data by randomly removing RGB modality or depth modality input with a certain probability $p$.
    \item \textbf{MM Prompt}~\cite{lee2023multimodal}: Building upon the Data Aug method, this method introduces an additional learnable token to indicate the current combination type of input modalities. (e.g., RGB-only, Depth-only, or RGB-D)
    \item \textbf{Ours$^{\ast}$}: To utilize the ground truth depth image, we replace the DCM described in \Cref{sec:method} with the depth branch as shown in the gray box in \Cref{fig:overview_method}.
\end{itemize}

\begin{table}[htbp]
    \centering
    \resizebox{0.4\textwidth}{!}{%
        \begin{tabular}{c|c|cccc}
             \toprule
             Method&  Avg&  Object&  Spatial&  \begin{tabular}[c]{@{}c@{}}Long\\ Horizon\end{tabular} & Goal\\ 
             \midrule
             RGB-RF &  57.95\%&  86.20\%&  69.60\%&  24.20\%& 51.80\%\\
             RGB-D-RF &  61.25\%&  \textbf{88.80\%} &  69.80\%&  30.20\%& 56.20\%\\ 
             Data Aug  &  58.45\% &  76.80\% &  65.80\%  &  35.60\%                                                 & 55.60\% \\
             MM Prompt & 58.75\% & 65.40\% & \textbf{77.40\%} & 22.00\% & \textbf{70.20\%} \\
             Ours$^{\ast}$      &  \textbf{63.95\%}  &  83.00\% &  69.80\% &  \textbf{37.40\%}                                                 & 65.60\% \\
             \bottomrule
        \end{tabular}%
    }
    \caption{Preliminary experiment results. We record the success rates of different models under \textbf{RGB-D} input (except for RGB-RF, which only utilizes RGB input).}
    \label{tab:preliminary_experiment_results}
    \vspace{-0.5cm}
\end{table}

\Cref{tab:preliminary_experiment_results} presents success rates of different models on the LIBERO benchmark when provided with RGB-D input. 
Our method achieves the best overall average success rate of 63.95\%, nearly a 6\% improvement over the baseline RGB-RF model. This demonstrates the superior performance of our method in effectively utilizing complete RGB-D input.
Further analyzing the different task subsets, our model achieves the highest success rate of 37.4\% on the Long Horizon suite. This indicates that our method can effectively utilize depth information, exhibiting a clear advantage in complex scenarios that require long-term planning. 
While the Data Aug and MM Prompt methods perform well on certain subsets, they have lower overall average success rates compared to RGB-D-RF and our method. This is due to the increased learning difficulty from needing to handle different modality combinations.
Additionally, we observe that our model achieves better results compared to directly introducing depth in RGB-D-RF. 
This shows that including the codebook effectively reduces noise-related errors in depth features, enhancing their robustness.

\subsection{Main Experiment}
In this section, we verify the task success rate of the model when only RGB images are provided as input. 
In addition to comparing with the baselines mentioned in \Cref{sec:preliminary_experiment}, we also compare our results with two cross-modal knowledge distillation methods. These methods distill knowledge from depth features into RGB features, enhancing the spatial perception ability of the RGB model.
Specifically, we compare the following two cross-modal knowledge distillation methods:
\begin{itemize}
    \item \textbf{CRD}~\cite{tian2019contrastive}: This is a cross-modal knowledge distillation method based on contrastive learning loss. 
    In the RGB-D-RF model, we use the depth branch to extract depth features as the teacher model. Additionally, we introduce the CRD contrastive learning loss function during the training of the RGB-RF model.
    \item \textbf{CMKD}~\cite{yuhong-CMKD-ECCV2022}: This is a cross-modal knowledge distillation method based on mean squared error (MSE) loss. 
    Similarly, in the RGB-D-RF model, we use the depth branch to extract depth features as the teacher model. We also introduce the MSE loss function during the training of the RGB-RF model.
\end{itemize}

\begin{table}[htbp]
    \centering
    \resizebox{0.4\textwidth}{!}{%
        \begin{tabular}{c|c|cccc}
             \toprule
             Method&  Avg&  Object&  Spatial&  \begin{tabular}[c]{@{}c@{}}Long\\ Horizon\end{tabular} & Goal\\
             \midrule
             RGB-RF&  57.95\% &  \textbf{86.20\%} &  69.60\%  &  24.20\%                                                 & 51.80\% \\
             RGB-D-RF&  15.65\% &  6.60\% &  22.00\%  &  3.80\%                                                 & 30.20\% \\
             Data Aug&  54.50\% &  73.80\% &  57.20\%  &  28.00\%                                                 & 59.00\% \\
             MM Prompt&  48.90\% &  53.40\% &  62.80\%  &  14.60\%                                                 & 64.80\% \\
             CRD&  47.60\% &  62.40\% &  50.20\%  &  19.60\%                                                 & 58.20\% \\
             CMKD&  50.60\% &  63.80\% &  60.60\%   &  14.60\%                                                 & 63.40\% \\
             Ours&  \textbf{63.15\%} &  78.60\% &  \textbf{71.20\%}  &  \textbf{36.40\%}                                                 & 
             \textbf{66.40\%} \\
             \bottomrule
        \end{tabular}%
    }
    \caption{Main Experiment Results. We record the success rates of different models in different suites under \textbf{RGB-only} input, with each task being tested \textbf{50} times. Specifically, in the RGB-D-RF model here, we do not use the Depth branch.}
    \label{tab:main_experiment_results}
    \vspace{-0.5cm}
\end{table}

The experimental results are shown in \Cref{tab:main_experiment_results}. 
Comparing the results in \Cref{tab:preliminary_experiment_results} and \Cref{tab:main_experiment_results}, we observe that in the RGB-only case, the average task success rates for all models have declined to varying degrees.
Specifically, the RGB-D-RF model shows the largest performance drop in the RGB-only scene. This model relies heavily on depth information and loses much of its manipulation capability without it.
Although the data augmentation methods, Data Aug and MM Prompt, achieve higher average success rates (54.50\% and 48.90\%, respectively) than the RGB-D-RF model (15.65\%) under RGB-only scene, they are lower than the baseline RGB-RF model (57.95\%). 
This result indicates that although Data Aug and MM Prompt aim to improve generalization and handle missing modalities through data augmentation, they have not succeeded in enhancing RGB-only performance. This has led to an overall decline in performance.
The two cross-modal knowledge distillation methods, CRD and CMKD, perform moderately but do not significantly outperform the baseline RGB-RF model.
This may be because these methods focus too much on transferring information from depth features in the RGB-RF model to RGB features in the RGB-D-RF model. They neglect the specific requirements of the manipulation task, which leads to performance degradation.
In contrast, our method achieves an average success rate of 63.95\% with RGB-D input. The success rate decreases only slightly to 63.15\% with RGB-only input, showing the smallest drop.
This advantage will make our method more applicable in practical application scenarios.

\begin{figure}[thpb]
    \begin{center}
        \centering
        \resizebox{0.8\linewidth}{!}{\includegraphics{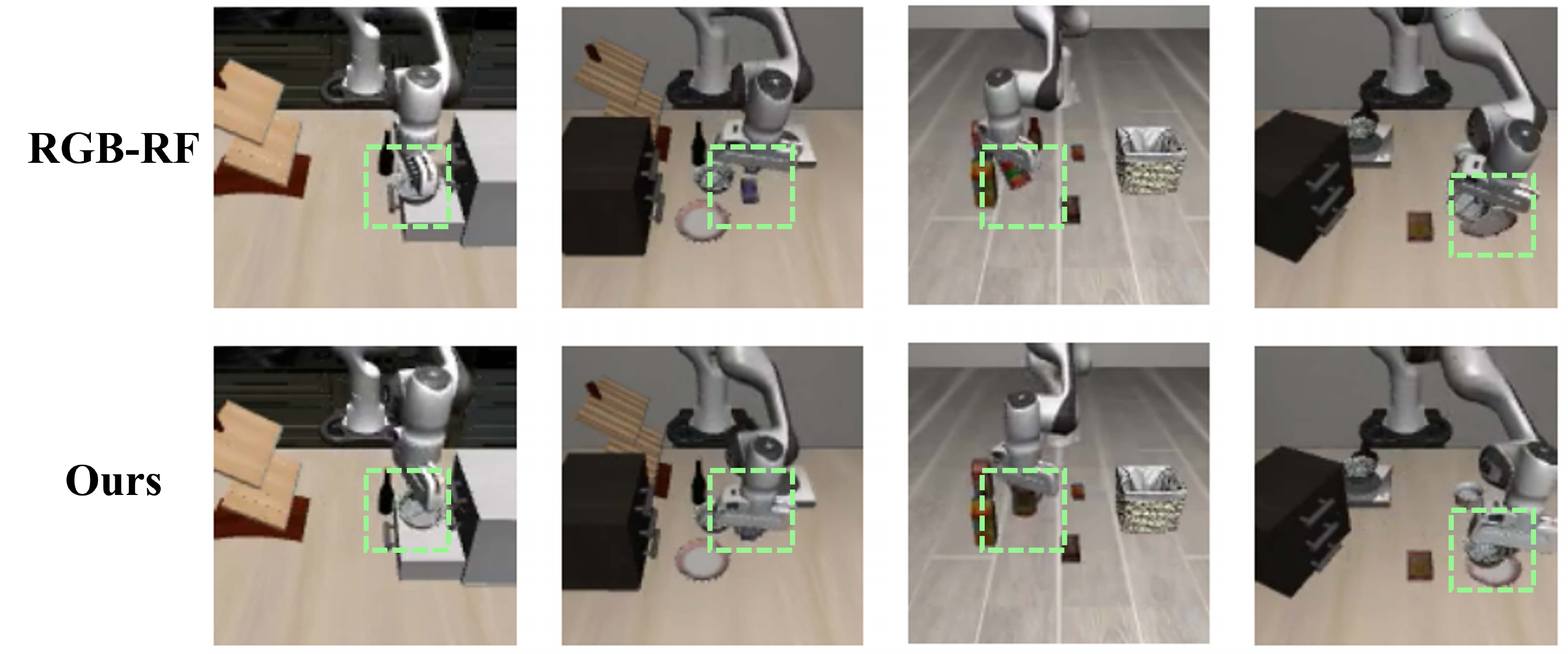}}
        \caption{Qualitative results. This figure illustrates the spatial position perception capability enabled by depth features. The top row shows the effects of the model without depth features. The bottom row shows the results of our method, which can complete the depth features using the depth completion module. }
        \label{fig:qualitative_results}
    \end{center}
    \vspace{-0.6cm}
\end{figure}

\begin{figure}[thpb]
    \begin{center}
        \centering
        \resizebox{0.8\linewidth}{!}{\includegraphics{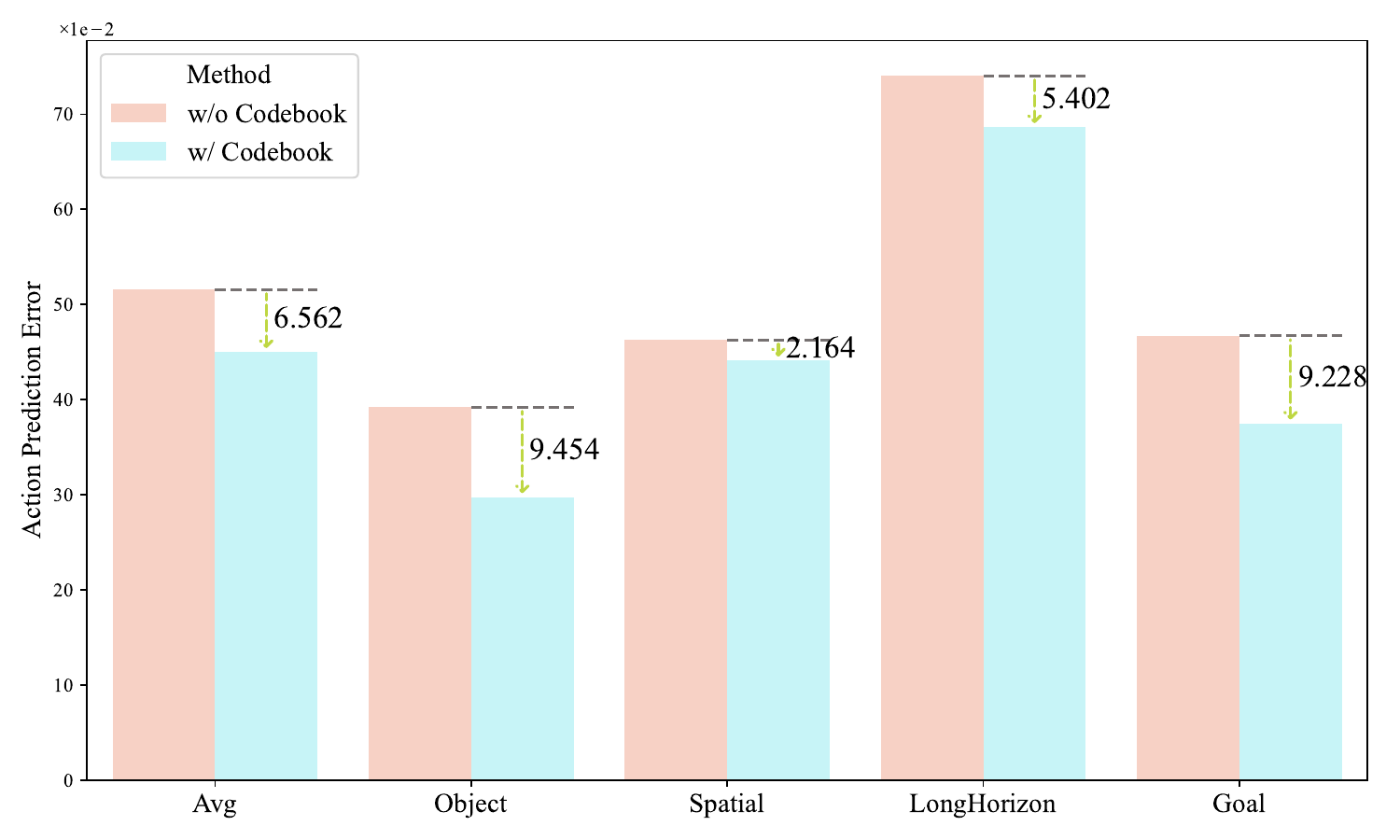}}
        \caption{Action Prediction Error. We evaluate the action sequences generated from predicted depth data against those derived from actual depth images and calculate the Euclidean distance between the actions at corresponding time steps.}
        \label{fig:vis_error}
        \vspace{-0.3cm}
    \end{center}
    \vspace{-0.3cm}
\end{figure}
\subsubsection*{\textbf{Qualitative Results}}
\Cref{fig:qualitative_results} provides visualized results to illustrate the role of depth information in decision-making. When the robotic arm approaches the target object, depth information plays a critical role. At this stage, accurately perceiving the 3D position and shape of the object is crucial for precise planning. Lacking depth information can lead to a biased understanding of the object's spatial position, affecting the accuracy of action planning.

\subsubsection*{\textbf{Prediction Error Analysis}}
To verify the role of the Depth-Aware Codebook in our framework, we compare the differences between the action sequences using different predicted depth methods and the action sequences planned using real-depth images. 
As shown in \Cref{fig:vis_error}, when using our proposed Depth-Aware Codebook, the error is smaller than when not using the Codebook. This result verifies that the Depth-Aware Codebook can effectively improve the quality of depth prediction, generating more accurate and higher-quality depth estimates. 
With optimized depth representation, our method maintains decision-making and control precision close to that with full RGB-D input, even without real depth data.

\subsubsection*{\textbf{Ablation Study}}

\begin{table}[htbp]
\centering
    \resizebox{0.4\textwidth}{!}{%
        \begin{tabular}{cc|c|cccc}
        \toprule
        \begin{tabular}[c]{@{}c@{}}DCM\end{tabular} & \begin{tabular}[c]{@{}c@{}}DAC\end{tabular}  & Avg     & Object  & Spatial & \begin{tabular}[c]{@{}c@{}}Long\\ Horizon\end{tabular} & Goal    \\
        \midrule
        $\times$   & $\times$   & 36.95\% & 33.80\% & 42.20\% & 14.20\% & 57.60\% \\
        \checkmark & $\times$   & 60.20\% & 87.00\% & 67.00\% & 34.40\% & 52.40\% \\
        \checkmark & \checkmark & 63.15\% & 78.60\% & 71.20\% & 36.40\% & 66.40\% \\
        \bottomrule
        \end{tabular}%
    }
    \caption{Ablation Experiment Results. $\times$ indicates not using that module, while \checkmark indicates using that module. In particular, we use an MLP to replace the Depth Completion Module if it is not being used. }
    \label{tab:ablation_study}
    \vspace{-0.4cm}
\end{table}

We conducted an ablation study on our model, with results shown in \Cref{tab:ablation_study}.  
The results show that the DCM significantly improved performance. It notably enhanced the model's results in the Object, Spatial, and Long Horizon suites.
The DAC mainly enhances the model's performance on the Goal test suite, while also bringing some improvements on the Spatial and Long Horizon suites. 
Notably, adding the Depth-Aware Codebook led to a performance drop in the Object suite. This result aligns with previous experiments in \Cref{tab:preliminary_experiment_results} and \Cref{tab:main_experiment_results} and may be related to the suite's characteristics.

\section{REAL WORLD EXPERIMENTS}

\begin{figure}[htpb]
    \begin{center}
        \centering
        \resizebox{0.8\linewidth}{!}{\includegraphics{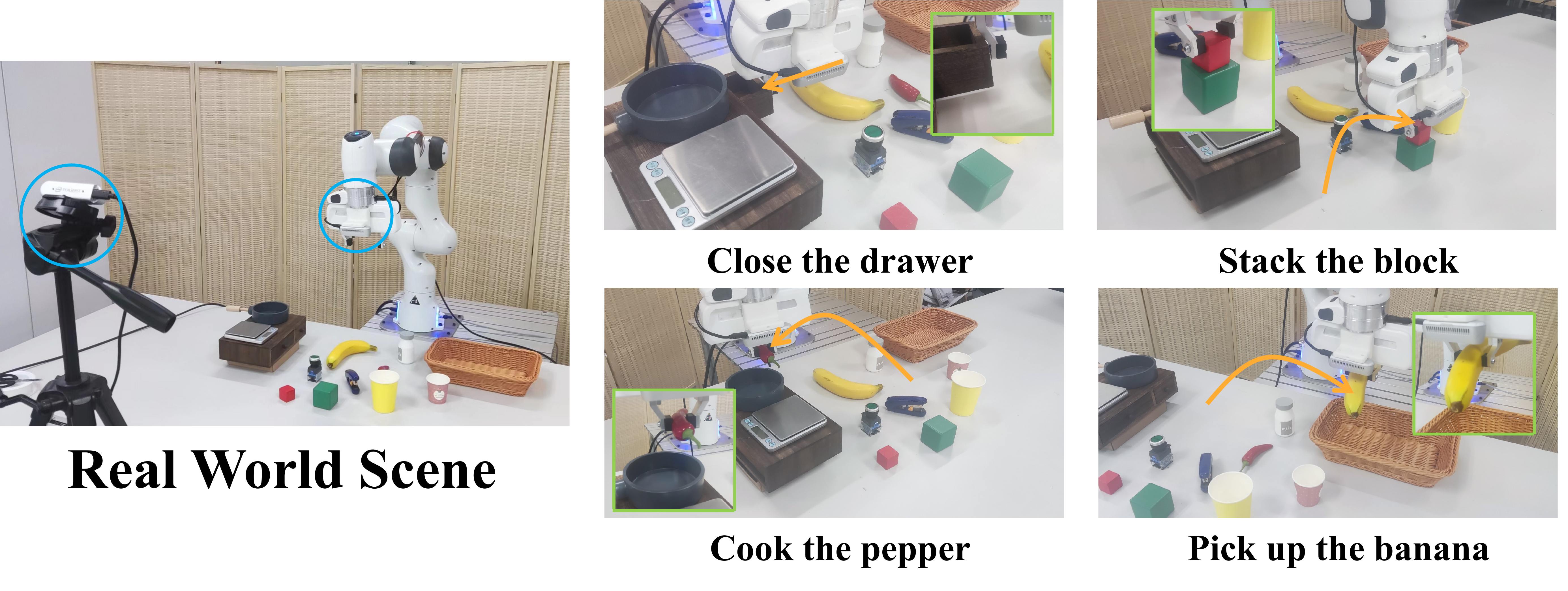}}
        \caption{Illustration of the real-world scene and the four tasks evaluated in our experiments.}
        \label{fig:real_demo}
    \end{center}
    \vspace{-0.7cm}
\end{figure}

\begin{table}[htpb]
\centering
    \resizebox{0.4\textwidth}{!}{%
        \begin{tabular}{c|c|cccc}
        \toprule
        Method  & Modality & Close drawer & \begin{tabular}[c]{@{}c@{}}Pick up \\ the banana\end{tabular} & \begin{tabular}[c]{@{}c@{}}Pick up \\ the pepper\end{tabular} & \begin{tabular}[c]{@{}c@{}}Stack \\ the block\end{tabular} \\
        \midrule
        RGB-RF  & RGB      & 13.33\%      & 26.67\%                                                       & 10.00\%                                                       & 20.00\%                                                    \\
        RGB-D-RF & RGB-D     & 63.33\%      & 83.33\%                                                       & 56.67\%                                                       & 53.33\%                                                    \\
        RGB-D-RF & RGB      & 3.33\%       & 20.00\%                                                       & 0.00\%                                                        & 6.67\%                                                     \\
        Ours    & RGB-D     & 73.33\%      & 76.67\%                                                       & 66.67\%                                                       & 73.33\%                                                    \\
        Ours    & RGB      & 66.67\%      & 80.00\%                                                       & 63.33\%                                                       & 73.33\%                                                    \\
        \bottomrule
        \end{tabular}%
    }
    \caption{Real World Experiment Result. }
    \label{tab:real_experiment_results}
    \vspace{-0.5cm}
\end{table}
To further verify the effectiveness of the proposed model, we conduct tests in the real world. The experimental setup utilized a Franka Emika Panda robotic arm equipped with two Intel D435 series cameras. 
We design a complex environment containing four different tasks. 
The third-person camera is positioned in front of the robotic arm. This setup makes it difficult for the model to determine the arm's position relative to the target object from texture information.
We collect 20 high-quality trajectories for each task using a SpaceMouse. These trajectories are then used to train a single multi-task model.
After training, the model is deployed on a host equipped with an NVIDIA 3090 GPU, and real-time inference is performed at a rate of 15 Hz. 
For each task,  we conduct 30 repeated tests to evaluate the model's performance in the real-world environment. 

The success rates are recorded in \Cref{tab:real_experiment_results}. The experimental results further validate the effectiveness of our method. 
Our method performs excellently with complete RGB-D input. More importantly, it also maintains outstanding performance even with just RGB input.
More details can be found in the \href{https://gewu-lab.github.io/DepthHelps-IROS2024}{supplementary video}.

\section{CONCLUSIONS}
 
In this paper, we propose the Depth Information Injection (${\rm DI}^2$) framework. It enhances the performance of pre-trained robot manipulation models that rely solely on RGB inputs by leveraging minimal aligned RGB-D trajectory data.
Our framework centers around two primary modules. 
The ${\rm DI}^2$ framework achieved better results in the LIBERO benchmark. Further, the results of the real-world experiments demonstrate the reliability and applicability of our method in practical application scenarios.

\section{ACKNOWLEDGEMENT}

This work is supported by the National Natural Science Foundation of China (NO.62106272), Shanghai AI Laboratory, National Key R\&D Program of China (2022ZD0160101), the National Natural Science Foundation of China (62376222), and Young Elite Scientists Sponsorship Program by CAST (2023QNRC001).

\bibliographystyle{IEEEtran}
\bibliography{IEEEabrv,mycite}

\end{document}